%% file: neurips_2026.tex
\documentclass{article}

\PassOptionsToPackage{numbers, compress}{natbib}
\usepackage[main, final]{neurips_2026}

\usepackage[utf8]{inputenc} 
\usepackage[T1]{fontenc}    
\usepackage{hyperref}       
\usepackage{url}            
\usepackage{booktabs}       
\usepackage{amsfonts}       
\usepackage{nicefrac}       
\usepackage{microtype}      
\usepackage{xcolor}

\usepackage{algorithmic}

\usepackage{xspace}
\usepackage{threeparttable}
\usepackage{multirow}
\usepackage{rotating}
\usepackage{makecell}
\usepackage{amsthm}

\usepackage{amsfonts}
\usepackage{flushend}
\usepackage{color}
\usepackage{bm}
\usepackage{subfigure}
\usepackage{bbm}
\usepackage{subcaption}
\usepackage{booktabs}
\usepackage{wrapfig}

\newcommand{\eat}[1]{}

\usepackage{enumitem}
\usepackage{pifont}
\usepackage{amsmath} 
\usepackage{algorithm}
\usepackage{array}
\usepackage{tcolorbox}
\usepackage[table]{xcolor}

\usepackage{textcomp}
\usepackage{stfloats}
\usepackage{url}
\usepackage{verbatim}
\usepackage{graphicx}

\newcommand{\ourmethod}{\textsc{AtlasULP}\xspace}

\usepackage{soul}

\newcommand{\tra}[1]{%
  {\sethlcolor{blue!20}\hl{#1}}%
}

\newcommand{\tes}[1]{%
  {\sethlcolor{green!20}\hl{#1}}%
}

\definecolor{win1}{rgb}{ .85,  .15,  .15 }
\definecolor{win2}{rgb}{ .439,  .678,  .278}
\definecolor{win3}{rgb}{ .267,  .447,  .769}

\title{\ourmethod: Domain-aware Universal Link Prediction via Relation Atlas}

\author{
Yujing Liu$^{1}$,\quad
Yixin Liu$^{1}$,\quad
Yu Zheng$^{1}$,\quad
Lianhua Chi$^{2}$ \\
\textbf{Alan Wee-Chung Liew$^{1}$,\quad
Heng Tao Shen$^{3}$,\quad
Shirui Pan$^{1}$}\thanks{Corresponding author.} \\[4pt]
$^{1}$School of Information and Communication Technology, Griffith University, Australia \\
$^{2}$Department of Computer Science and Information Technology, La Trobe University, Australia \\
$^{3}$School of Computer Science and Technology, Tongji University, China \\[4pt]
\texttt{yujing.liu@griffithuni.edu.au} \quad
\texttt{s.pan@griffith.edu.au}
}

\begin{document}

\maketitle

\begin{abstract}
Link prediction (LP) is a widely applied task in graph learning. Conventional LP approaches typically follow a dataset-specific paradigm, requiring independent training for each graph and incurring high computational and maintenance costs in large-scale applications. Motivated by these limitations, recent work explores Universal Link Prediction (ULP), aiming to enable training-free inference on arbitrary unseen graphs. However, existing methods primarily rely on subgraph sampling strategies to construct universal representations, which are unable to capture higher-order information due to the exponential growth of higher-order neighborhoods. As a result, they often under-perform on graphs where long-range signals are essential, such as biological networks. Moreover, their prediction models are typically domain-agnostic and lack the ability to adapt to different graph domains, leading to limited generalization.
To address these challenges, we propose an Atlas-based Universal Link Prediction framework (\ourmethod), which introduces the relation atlas to capture structural associations between node pairs from both global and local perspectives, enabling a comprehensive characterization of connectivity signals across diverse graph domains. Building upon this representation, we develop a domain-guided prompting ULP model, which generates domain-aware prompt tokens from contextual structures and performs prompt-guided in-context prediction for adaptive link inference. Extensive experiments demonstrate that \ourmethod consistently outperforms state-of-the-art methods across diverse graph domains.
\end{abstract}

\section{Introduction}
\input{backup/intro}

\input{backup/rw}

\section{Method}
\input{backup/method}

\section{Experiments}
\input{backup/exp}

\section{Conclusion}
\input{backup/conclusion}

\clearpage

\section*{Acknowledgments and Disclosures}

\noindent\textbf{Funding.}
The work of S. Pan was partially supported by the Australian Research Council (ARC) under Grant Nos. DP240101547 and FT210100097. The work of Y. Liu was partially supported by the ARC under Grant No. DE260101172.

\noindent\textbf{Competing Interests.}
The authors declare no competing interests.

\bibliography{reference}
\bibliographystyle{abbrvnat}

\clearpage

\appendix
\input{backup/appendix}

\clearpage
\newpage
\input{checklist.tex}

\end{document}

%% file: backup/intro.tex
Link prediction (LP) aims to infer missing or potential interactions between entities in graph data, and is one of the most fundamental tasks in graph learning. With the widespread adoption of graph-structured data in real-world scenarios such as social networks~\cite{amran2026link,jaskari2025temporal}, biological networks~\cite{horecka2026risk,nicolle2026pagan}, and information networks~\cite{pan2024unifying,robinson2024relational}, LP techniques have been extensively applied to address practical problems such as friend recommendation~\cite{wu2021self,wu2020diffnet}, protein–protein interaction prediction~\cite{reau2023deeprank,sledzieski2021d}, and knowledge graph completion~\cite{galkin2021nodepiece,wang2022simkgc}.
Despite the remarkable success of Graph Neural Networks (GNNs)~\cite{li2023evaluating,liu2025hierarchical,zhu2021neural} and various heuristic algorithms~\cite{fang2024link,gui2024link} in LP, most existing approaches follow a dataset-specific paradigm, requiring independent model training or heuristic selection for new graphs. This results in heavy computational and operational costs, hindering their large-scale deployment in industrial settings and their applicability in real-world applications with limited training data. 

To overcome the above limitations, a promising direction is to develop universal models~\cite{liu2026real,li2026relationalAD,liu2026rethinking} that enable training-free LP on unseen graphs without dataset-specific adaptation. 
Distinct from other graph learning tasks such as node classification~\cite{liu2025noisy} and graph classification~\cite{ju2025cluster}, LP is inherently more dependent on topological structures than node attributes, as it aims to infer missing links based on structural connectivity patterns. Moreover, in several real-world scenarios (e.g., protein-protein interaction networks), graph data are often unattributed, where only topological information is available. 
As a result, graph foundation models designed for other graph learning tasks often exhibit limited performance or even become inapplicable in LP due to their reliance on node attributes. 
In this context, Dong et al. conducted a pioneering exploration into universal link prediction (ULP), aiming to train models on source graphs from multiple domains and enable effective link prediction on unseen target graphs without fine-tuning. To this end, they proposed UniLP~\cite{dong2024universal}, a framework that extends Double Radius Node Labeling (DRNL)~\cite{zhang2018link} to construct universal structural representations and incorporates an attention-based in-context learning model to predict links on unseen graphs.

\begin{wrapfigure}{r}{0.35\textwidth}
 \centering
  \includegraphics[width=0.35\textwidth]{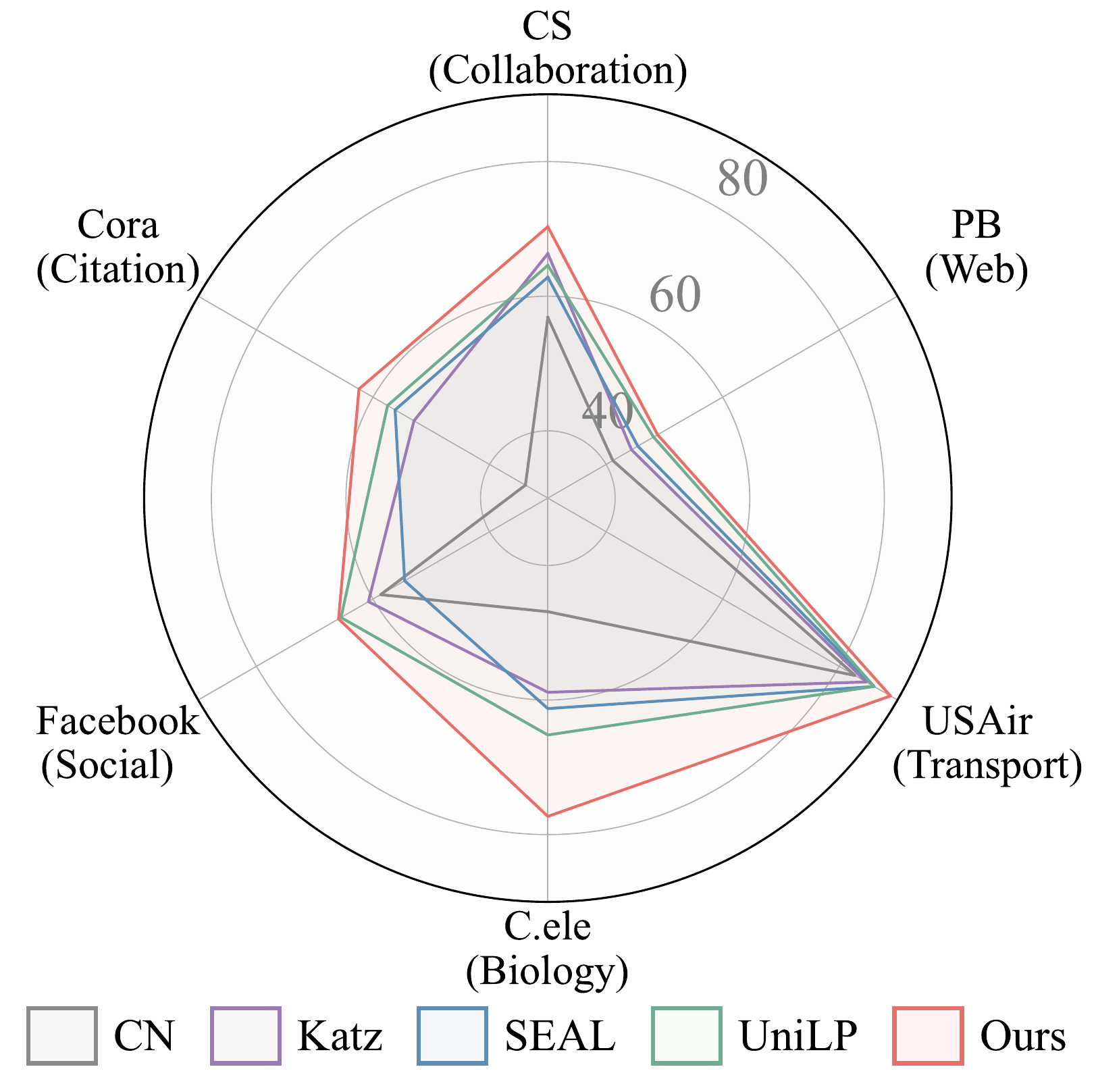}
  \vspace{-6mm}
  \caption{Generalization across diverse graph domains.}
  \label{fig:intro_radar}
  \vspace{-2mm}
\end{wrapfigure}

Despite its pioneering efforts, UniLP still exhibits limited generalizable prediction capability across different domains, as it struggles to tackle two fundamental challenges in ULP. The first would be: 
\textit{\textbf{C1 - How to construct a universal representation that can capture connectivity patterns across different graph domains?}} Due to domain heterogeneity, graphs from different domains exhibit distinct structural patterns and hence rely on different structural cues for link prediction~\cite{clauset2008hierarchical,liben2003link,lu2011link}. For example, in social networks, links exhibit strong local clustering, well predicted by local patterns such as common neighbors~\cite{liben2003link}. In contrast, in citation networks, local patterns are often ineffective due to the scarcity of local topology. In these cases, global approaches such as the Katz index can better characterize higher-order path dependencies~\cite{clauset2008hierarchical,lu2011link}. Such significant diversity underscores the need for comprehensive representations that can capture multi-scale structural information for ULP. Nevertheless, UniLP only leverages DRNL on local ego-subgraphs as the universal representation, which emphasizes the importance of 1–2-hop neighborhoods but overlooks higher-order structural signals, leading to suboptimal performance on graphs where long-range structural dependencies are critical, such as citation (Cora) and biological (Celegans) networks (Fig.~\ref{fig:intro_radar}). 

While the representations serve as a unified input that provides rich structural information, the universal model still plays the critical role of learning both shared knowledge and domain-specific variations across different graph domains. This further motivates the second challenge: 
\textit{\textbf{C2 - How to design universal models that can adaptively identify domain-specific connectivity signals?}}
Since different graph domains exhibit distinct connectivity patterns, a comprehensive universal representation inevitably contains both useful and redundant structural signals depending on the target domain. Therefore, an ideal ULP model should go beyond a static prediction paradigm, possess the ability to perceive the connectivity patterns of the target domain. Ideally, it should further leverage structural priors learned from pre-training to capture discriminative structural signals from universal representations, thereby enabling cross-domain transfer. However, UniLP only partially fulfills this ideal, as its adaptive mechanism is confined to leveraging few-shot context samples for prediction, while failing to model domain-specific structural patterns with dynamic adaptation.

In this work, we tackle the above two challenges by proposing \ourmethod, a novel ULP framework. 
Our theme is to learn a comprehensive representation that captures multi-scale structural information, and couple it with a domain-adaptive inference model for ULP. 
More specifically, to address \textbf{\textit{C1}}, we introduce Relation Atlas (RA), a new pair-level universal representation tailored for ULP, encoding multi-scale structural information from both global and local perspectives. At the global level, we leverage multi-order Chebyshev recursion to capture multi-order path responses, thereby modeling long-range topological dependencies. At the local level, we characterize fine-grained structural relationships using statistical measures that capture local connectivity patterns and structural role similarity, complementing global representations. 
Building upon RA, to handle \textbf{\textit{C2}}, we propose a domain-guided prompting ULP model, consisting of two modules: domain-aware prompt learning and prompt-guided in-context prediction. Conditioned on the RA representations of the target dataset, the former constructs domain-specific structural prompts from a set of learnable structural probes, enabling the model to explicitly model domain-specific information. Powered by a Transformer backbone, the latter performs domain-adaptive link prediction by leveraging domain prompts and contextual information, thereby facilitating knowledge transfer across diverse graph domains. To sum up, the main contributions of this paper are three-fold:

\begin{itemize}[leftmargin=*]
\item[$\bullet$] \textbf{Unified Representation.} We rethink the construction of universal representations in existing LP methods and identify their limitations, motivating a relation atlas (RA) that captures both global and local structural associations between node pairs.

\item[$\bullet$] \textbf{Universal Model.} We propose \ourmethod, an Atlas-grounded universal LP framework that builds upon RA to enable effective cross-domain generalization via domain-aware prompt learning and prompt-guided in-context link prediction.

\item[$\bullet$] \textbf{Experiments.} Extensive experimental results on graph datasets from six real-world domains validate the effectiveness, transferability, and scalability of \ourmethod.
\end{itemize}

%% file: backup/rw.tex
\section{Related Work}
\subsection{Link Prediction.} 
With the increasing prevalence of graph-structured data, graph learning techniques~\cite{zhao2026fedcigar,pan2026camera,liu2026pasta} have been widely applied to various real-world problems. Among them, link prediction (LP) is a fundamental problem in graph mining. Early approaches mainly rely on heuristic measures that quantify node similarity based on graph topology. Representative methods include local heuristics such as Common Neighbors (CN)~\cite{liben2003link}, Adamic-Adar (AA)~\cite{adamic2003friends}, and Resource Allocation (RA)~\cite{zhou2009predicting}, as well as global measures such as the Katz index~\cite{katz1953new}, which captures higher-order path dependencies. While these methods are computationally efficient and interpretable, they rely on predefined rules and lack adaptability.

More recently, learning-based approaches have achieved significant success by automatically learning structural representations. Early embedding-based methods, such as DeepWalk~\cite{perozzi2014deepwalk} and node2vec~\cite{grover2016node2vec}, learn node representations via random walks, while matrix factorization methods~\cite{menon2011link} model latent interactions. Graph neural networks (GNNs) further advance LP by learning expressive node representations through message passing~\cite{kipf2016semi}. Variants such as SEAL~\cite{zhang2018link} enhance performance by leveraging enclosing subgraphs and structural labeling (e.g., DRNL) to encode local topology, while recent methods (e.g., ELPH~\cite{chamberlain2022graph}, NCNC~\cite{wang2023neural}, MPLP~\cite{dong2024pure}) improve scalability and structural modeling. Despite their strong performance, most learning-based methods follow a dataset-specific training paradigm, requiring separate training for each graph, which limits scalability and generalization to unseen domains.

\subsection{Universal Link Prediction.} 
To overcome the limitations of the dataset-specific paradigm, recent efforts have begun to explore foundation models for graph learning, aiming to generalize across diverse graph domains. Graph Foundation Models (GFMs), such as OFA~\cite{liu2024one} and PRODIGY~\cite{huang2023prodigy}, have shown promising results in zero-shot node and graph classification by leveraging node attributes and prompt-based task adaptation. However, their effectiveness in link prediction remains limited, as LP primarily depends on structural connectivity rather than semantic attributes, making attribute-centric approaches less suitable for unattributed graphs.

Consequently, Universal Link Prediction (ULP) has emerged to explicitly learn transferable topological patterns. UniLP~\cite{dong2024universal} represents a pioneering effort in this direction. It extends Double Radius Node Labeling to construct universal structural representations and employs an attention-based in-context learning mechanism with target graph exemplars to enable training-free inference. Despite its promising results, UniLP still faces notable limitations. First, its reliance on local ego-subgraphs restricts the modeling of higher-order structural dependencies (akin to those captured by Katz or NBFNet). Second, its adaptation mechanism utilizes a domain-agnostic matching function. Since different domains exhibit highly heterogeneous connectivity priors (e.g., homophily in social networks versus structural equivalence in biological networks), such a design may limit its ability to adaptively capture domain-specific patterns.

\section{Preliminary}
\textbf{Notations.}
Let $\mathcal{G}=(\mathcal{V},\mathcal{E})$ denote a graph, where $\mathcal{V}=\{v_1,\cdots,v_n\}$ is the set of $n$ nodes and $\mathcal{E}\subseteq \mathcal{V}\times\mathcal{V}$ represents the set of observed edges. The graph topology is encoded by an adjacency matrix $\mathbf{A}\in\mathbb{R}^{n\times n}$, where $a_{ij}=1$ if $(v_i,v_j)\in\mathcal{E}$ and $a_{ij}=0$ otherwise.
Given a edge $e = (v_i, v_j)$, the link prediction (LP) task aims to estimate the likelihood of an edge existing between them. We denote $y_{e}\in\{0,1\}$ as the link label, where $y_{e}=a_{ij}$ indicates whether the edge exists.

\textbf{Universal Link Prediction Problem.}
The goal of universal link prediction (ULP) is to learn a universal model $f_{\theta}(v_i, v_j)$ that can generalize across diverse graph domains. Formally, let $\mathcal{T}_{s}=\{\mathcal{G}_{s}^{(1)}, \cdots, \mathcal{G}_{s}^{(N)}\}$ denote a collection of source-domain graphs used for training, and $\mathcal{T}_{t}=\{\mathcal{G}_{t}^{(1)}, \cdots, \mathcal{G}_{t}^{(N')}\}$ denote unseen target-domain graphs. The objective is to learn a ULP parameterized by $\theta$ from source-domain graphs, which can be directly applied to unseen target graphs without further fine-tuning. 
During inference on target graphs, observed edges naturally serve as contextual references, providing in-context information without additional cost. The model leverages such structural context to guide link inference. The final objective is to correctly rank missing edges higher than non-edge node pairs in the target graph.

%% file: backup/method.tex
\begin{figure*}[t!]
    \centering
    \includegraphics[width=\textwidth,
        ]{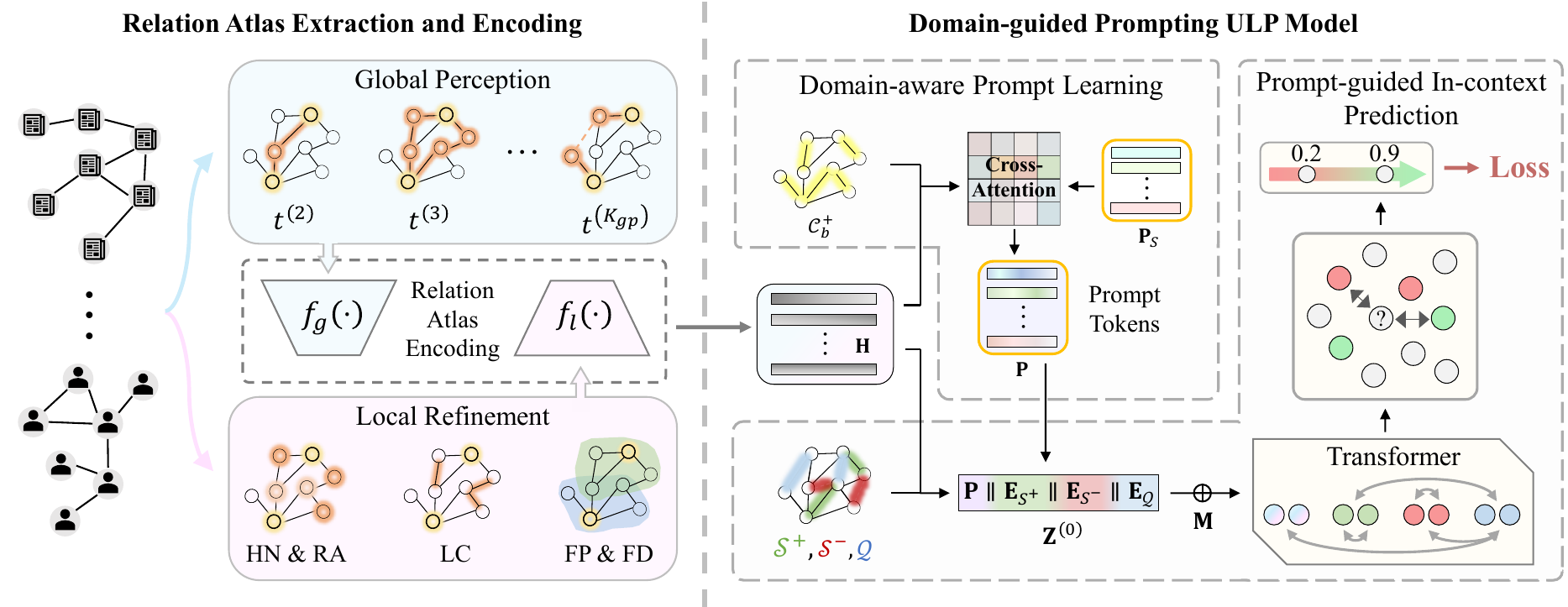}
    \caption{The framework of the proposed \textbf{\ourmethod}.}\label{fig:framework} 
\end{figure*}

In this section, we present \ourmethod, an relation \textbf{Atlas}-based \textbf{U}niversal \textbf{L}ink \textbf{P}rediction framework~(Fig.~\ref{fig:framework}). \ourmethod consists of two components: Relation Atlas extraction and encoding, which captures and encodes multi-scale topological associations between node pairs into a universal LP representation; and a domain-guided prompting ULP model, which performs domain-adaptive link inference via domain-aware prompt learning and prompt-guided in-context prediction.

\subsection{Relation Atlas Extraction and Encoding}

To develop a generalizable ULP framework, a key prerequisite is to map graph data from different domains into a unified and comparable representation space. 
As predicting links across graph domains may rely on different structural cues, a universal representation should integrate multi-scale topology to cover comprehensive connectivity patterns. To this end, we propose Relation Atlas (RA), a node-pair-level unified representation that explicitly models these structural priors through both \textit{global perception} and \textit{local refinement}.

\subsubsection{Relation Atlas Extraction}
\textbf{Global Perception.} As the main component of RA, the \textit{global perception} characterizes global structural relationships by capturing multi-order path response intensities between node pairs. Such higher-order signals are essential in sparse graphs such as citation networks, where many linked paper pairs share no common neighbors due to sparse citation structures and are only connected via multi-hop citation chains.
While several traditional heuristics (e.g., Katz and PageRank) are designed for capturing high-order structural information, they aggregate multi-order topology via predefined schemes, which may blur contributions across scales and limit their ability to distinguish heterogeneous higher-order dependencies across graph domains. To address this, we propose multi-order path response modeling, which leverages Chebyshev Polynomial Recursion to explicitly encode order-wise response strengths. Given the normalized adjacency matrix $\tilde{\mathbf{A}} = \mathbf{D}^{-1/2}\mathbf{A}\mathbf{D}^{-1/2}$, we compute the $a$-th order polynomial matrix $\mathbf{T}_a$ recursively:

\begin{equation}
\mathbf{T}_a = 2\widetilde{\mathbf{A}}\mathbf{T}_{a-1} - \mathbf{T}_{a-2},
\label{eq1}
\end{equation}

where $\mathbf{T}_0 = \mathbf{I}$ and $\mathbf{T}_1 = \widetilde{\mathbf{A}}$. The response strength for a node pair $(v_i, v_j)$ at order $a$ can be extracted from the corresponding matrix element $t_{i,j}^{(a)} = \left[ \mathbf{T}_a \right]_{i,j}$. Intuitively, $t_{i,j}^{(a)}$ reflects the degree of path-based association between $v_i$ and $v_j$ at $a$ hops, capturing scale-specific dependencies in the global topology.
This strategy effectively suppresses backtrack signals through differential cancellation between recursive terms, mitigating the homogenization of high-order signals common in $\tilde{\mathbf{A}}^k$-based methods~\cite{krzakala2013spectral}. We then concatenate the responses across $K_{gp}$ orders to obtain the global perception representation of node pair $(v_i, v_j)$:

\begin{equation}
\mathbf{h}_{global}^{(i,j)} = \left[ t_{i,j}^{(1)} \parallel t_{i,j}^{(2)} \parallel \cdots \parallel t_{i,j}^{(K_{gp})} \right],
\label{eq2}
\end{equation}

where $K_{gp}$ denotes the maximum path order considered in global perception. This yields a comprehensive multi-scale characterization of global structural relationships, providing clear and informative topological signals for downstream general models.

\textbf{Local Refinement.} While global perception captures long-range structural signals, it is less effective in modeling fine-grained local interactions, which is also crucial in many graphs. To complement this, we introduce \textit{local refinement}, another key component of RA that captures $K_{lr}$-hop local structural patterns using lightweight statistical indicators, avoiding costly subgraph extraction, where $K_{lr}$ denotes the maximum hop considered in local refinement.
Specifically, we characterize local structural patterns from two complementary perspectives: \textit{local connectivity patterns} and \textit{structural role similarity}. Detailed mathematical formulations are deferred to Appendix~\ref{app: local}.

\textit{Local connectivity patterns} serve as strong link indicators in many graphs~\cite{liben2003link,lu2011link}, where node pairs with more shared neighbors (e.g., mutual friends in social networks) are more likely to form links. In this paper, we characterize local connectivity patterns using three measures: Hierarchical Neighborhood Overlap ($\mathbf{s}_{\text{HN}}$), Resource-Allocated Overlap ($\mathbf{s}_{\text{RA}}$), and Local Connectivity Density ($\mathbf{s}_{\text{LC}}$). $\mathbf{s}_{\text{HN}}$ captures the quantity of neighbor overlap across multi-hop structural scales. $\mathbf{s}_{\text{RA}}$ introduces degree information to differentiate the importance of common neighbors, acknowledging that low-degree shared neighbors often indicate tighter connections than high-degree hubs. $\mathbf{s}_{\text{LC}}$ further quantifies the edge connection strength between cross-level neighborhoods to capture topological compactness that cannot be expressed by overlap alone. Through these three metrics, we obtain a detailed characterization of local subgraph structure. Notably, as all metrics are computed via sparse matrix operations without subgraph extraction, the design is computationally efficient. 

\textit{Structural role similarity} characterizes the node pairs by reflecting each node’s position in local topology (e.g., hubs or peripheral nodes) and serves as an important signal for predicting connectivity behavior in real-world graphs~\cite{ribeiro2017struc2vec,henderson2012rolx}. For example, prominent hubs (e.g., celebrities in social networks) inherently possess a higher propensity to interconnect, whereas peripheral nodes are usually confined to isolated cliques. We characterize structural role similarity using Frontier Size Product ($\mathbf{s}_{\text{FP}}$) and Frontier Size Difference ($\mathbf{s}_{\text{FD}}$). $\mathbf{s}_{\text{FP}}$ measures the product of frontier set sizes across different orders, while $\mathbf{s}_{\text{FD}}$ computes their absolute discrepancy. Jointly modeling these metrics provides a comprehensive view of structural roles. For instance, a small $\mathbf{s}_{\text{FD}}$ coupled with a high $\mathbf{s}_{\text{FP}}$ suggests symmetric hub-like nodes, which typically exhibit higher link formation likelihood than symmetric peripheral ones.

By combining all five structural metrics, we obtain a unified representation of local refinement representation, which is defined as follows:

\begin{equation}
\mathbf{h}_{local}^{(i,j)} = \left[ \mathbf{s}_{\text{HN}}^{(i,j)} \parallel \mathbf{s}_{\text{RA}}^{(i,j)} \parallel \mathbf{s}_{\text{LC}}^{(i,j)} \parallel \mathbf{s}_{\text{FP}}^{(i,j)} \parallel \mathbf{s}_{\text{FD}}^{(i,j)} \right].
\label{eq3}
\end{equation}

Finally, by combining the global perception and local refinement representations, we obtain a comprehensive modeling of multi-scale topological signals, forming an effective universal LP representation.

\subsubsection{Relation Atlas Encoding}
After constructing the global perception representation $\mathbf{h}_{global}^{(i,j)}$ and the local refinement representation $\mathbf{h}_{local}^{(i,j)}$, we project them into a continuous embedding space. Since global and local signals capture different topological scales and exhibit distinct feature distributions, directly fusing them may cause one signal to dominate the other or dilute its contribution. Therefore, we employ a dual-stream encoder that enables separate non-linear transformations for each component. Specifically, the global perception and local refinement representations are separately encoded and then concatenated to obtain the final embedding for any given edge $e = (v_i, v_j)$:
\begin{equation}
\mathbf{h}_{e} = \left[ f_{g}(\mathbf{h}_{global}^{(i,j)}) \parallel f_{l}(\mathbf{h}_{local}^{(i,j)}) \right],
\label{eq4}
\end{equation}
where $f_{g}(\cdot)$ and $f_{l}(\cdot)$ denote the global perception encoder and local refinement encoder, respectively. This comprehensive representation $\mathbf{h}_{e}$ captures multi-scale topological signals and serves as the foundation for the subsequent training-free inference.

\subsection{Domain-guided Prompting ULP Model}

While the aforementioned universal representation comprehensively captures pairwise structural correlations, real-world graph domains often exhibit heterogeneous connectivity patterns. Consequently, structural signals that are highly predictive in one domain may act as redundant noise in another. To perform effective training-free inference, an ideal universal model must adaptively distill valid link-predictive signals from the universal representation while suppressing domain-irrelevant interference. To this end, we design a domain-guided prompting ULP model, consisting of Domain-aware Prompt Learning and prompt-guided in-context prediction.

\subsubsection{Domain-aware Prompt Learning}
To explicitly model the connection patterns of target domain, we propose a perceiver-based domain prompt generator that extracts $K$ domain-aware prompt tokens from context samples. These tokens, carrying domain-level information, can provide domain-related guidance for inference.

The domain-aware prompt learning module takes edge representations as input to construct prompt tokens. However, given the diverse scale of potential target graphs, using all edges to construct prompt tokens may incur prohibitive computational and memory costs for some large-scale data. In the other hand, random sampling may fail to capture rare structural patterns (e.g., sparse connections involving low-degree nodes). To address this, we adopt a lightweight context selection strategy to select representative samples. Specifically, we rank candidate edges by the sum of endpoint degrees and apply uniform interval sampling to select $M$ background reference edges, i.e., $\mathcal{C}^+_{b} = { e_1, \cdots, e_M }$. Then, their representations $\mathbf{H}_{b} = [\mathbf{h}_{e_1}, \cdots, \mathbf{h}_{e_M}]^\top \in \mathbb{R}^{M \times d}$ can serve as the input of the prompt learning module. 
This strategy ensures balanced coverage from sparse regions to hub-dominated structures with low overhead, preserving the diversity of structural patterns in the context.

Based on the selected background reference edges, we design a perceiver-based domain-aware prompt generator that uses $K$ learnable structural probes $\mathbf{P} \in \mathbb{R}^{K \times d}$ to distill the topological patterns of target dataset via a cross-attention block where $\mathbf{H}_{b}$ serves as keys and values:

\begin{equation}
\mathbf{P}^{'} = \text{Softmax}\left( \frac{\mathbf{P} \mathbf{H}_{b}^\top}{\sqrt{d}} \right) \mathbf{H}_{b}.
\label{eq5}
\end{equation}

Subsequently, we feed the intermediate embedding $\mathbf{P}^{'}$ into a Feed-Forward Network (FFN) for non-linear mapping and refinement to obtain $K$ domain-aware prompt tokens:

\begin{equation}
\mathbf{P}_{domain} \in \mathbb{R}^{K\times d} = \text{FFN}(\text{LayerNorm}(\mathbf{P} + \mathbf{P}^{'})) + \mathbf{P}^{'}.
\label{eq6}
\end{equation}

Through this process, we compress domain-related background information into a fixed number of domain-level prompt tokens, mapping instance-level structural observations to domain-level summaries. These prompts are independent of specific edges, providing compact domain-aware signals for subsequent inference.

\subsubsection{Prompt-guided In-context Prediction}
After obtaining the domain-aware prompt tokens $\mathbf{P}_{domain}$, we perform prompt-guided in-context prediction to adaptively predict query edges under prompt conditioning. We use a Transformer as the in-context encoder, as its expressive power and scalability make it well suited for serving as a foundation model. 
We also construct a refined support set from the filtered background context to serve as a concrete reference of connected/disconnected examples. To ensure low inference overhead, we design a deterministic node pair sampling strategy to construct the context sample set. Given background reference edges $\mathcal{C}^+_{b}$ and negative pool $\mathcal{C}^-$ (globally sampled negatives), we assign each candidate edge $e=(v_i, v_j)$ a structural typicality score. The score is computed by weighting the 1-hop components of hierarchical neighborhood overlap and local connectivity density:

\begin{equation}
o(e) = \lambda_1 \cdot (\mathbf{f}_{v_i}^{(1)})^\top \mathbf{f}_{v_j}^{(1)} + \lambda_2 \cdot (\mathbf{f}_{v_i}^{(1)})^\top \mathbf{A} \mathbf{f}_{v_j}^{(1)} + \epsilon \cdot \text{idx}(e),
\label{eq7}
\end{equation}
where $\mathbf{f}_{v_i}^{(a)} \in \{0,1\}^n$ is the binary indicator vector of the $a$-hop frontier set of node $v_i$, $\lambda_1$ and $\lambda_2$ are weighting coefficients, and $\epsilon$ is a small tie-breaking regularizer based on the edge index. Subsequently, we sort $\mathcal{C}^+_{b}$ and $\mathcal{C}^-$ by $o(e)$ and apply uniform interval sampling to obtain support sets $\mathcal{S}^+$ and $\mathcal{S}^-$. This ensures that $\mathcal{S}^-$ covers hard negatives with high structural overlap but no observed links, improving boundary separation in in-context inference. 

Next, we feed domain prompts, context references, and query edges into the Transformer for in-context representation learning. Given the refined support set $\mathcal{S} = {\mathcal{S}^+, \mathcal{S}^-}$ and query set $\mathcal{Q}$, we first obtain their edge embeddings via the dual-stream encoder (Eq.~\eqref{eq4}). We then construct the input sequence of Transformer consisting of domain prompts, context references, and query samples, distinguished by role embeddings:

\begin{equation}
\mathbf{Z}^{(0)} = \left[ \mathbf{P}_{domain} \parallel \mathbf{E}_{\mathcal{S}^+} \parallel \mathbf{E}_{\mathcal{S}^-} \parallel \mathbf{E}_{\mathcal{Q}} \right],
\label{eq8}
\end{equation}

where learnable role embeddings are incorporated into each component, e.g., $\hat{\mathbf{P}}_{domain} = \mathbf{P}_{domain} + \mathbf{e}_{role}^{(p)}$, and $\mathbf{E}_{\mathcal{S}^+}$, $\mathbf{E}_{\mathcal{S}^-}$, $\mathbf{E}_{\mathcal{Q}}$ are obtained by adding corresponding role embeddings $\mathbf{e}_{role}^{(s+)}, \mathbf{e}_{role}^{(s-)}, \mathbf{e}_{role}^{(q)}$ to the dual-stream edge embeddings $\mathbf{H}$. To enforce the inductive setting in in-context inference, we design an attention mask $\mathbf{M} \in \mathbb{R}^{L_{seq} \times L_{seq}}$ to prevent information leakage among query samples, and feed the input sequence $\mathbf{Z}^{(0)}$ together with $\mathbf{M}$ into an $L$-layer Transformer encoder (see Appendix~\ref{app: mask} for details), yielding the final representations $\mathbf{Z}^{(L)} = \left[ \tilde{\mathbf{P}}_{domain} \parallel \tilde{\mathbf{E}}_{\mathcal{S}^+} \parallel \tilde{\mathbf{E}}_{\mathcal{S}^-} \parallel \tilde{\mathbf{E}}_{\mathcal{Q}} \right]$. The domain prompt tokens are then discarded, as they serve as contextual priors, and extract updated representations for positive supports $\tilde{\mathbf{E}}_{\mathcal{S}^+}$, negative supports $\tilde{\mathbf{E}}_{\mathcal{S}^-}$, and queries $\tilde{\mathbf{E}}_{\mathcal{Q}}$ from the final representations for link inference.

Subsequently, for each query edge ${e_i} \in \mathcal{Q}$, we evaluate its topological affinity by calculating its average similarity with the positive and negative support sets:

\begin{equation}
    s^{+}({e_i}) = \frac{1}{|\mathcal{S}^+|} \sum_{e^+ \in \mathcal{S}^+} \text{sim}(\textbf{z}_{e_i}, \textbf{z}_{e^+}) \quad \text{and} \quad s^{-}({e_i}) = \frac{1}{|\mathcal{S}^-|} \sum_{e^- \in \mathcal{S}^-} \text{sim}(\textbf{z}_{e_i}, \textbf{z}_{e^-}),
\label{eq9}
\end{equation}

where $\textbf{z}_{e_i}$, $\textbf{z}_{e^+}$, and $\textbf{z}_{e^-}$ are feature vectors extracted from the updated representations, and $\text{sim}(\cdot,\cdot)$ denotes cosine similarity in a linearly projected metric space. Based on the relative similarity, the final link probability of query edge $q$ is computed as:

\begin{equation}
\tilde{y}_{e_i} = \sigma \big( \tau \cdot ( s^{+}({e_i}) - s^{-}({e_i}) ) + b \big),
\label{eq10}
\end{equation}

where $\sigma(\cdot)$ is the sigmoid function, $\tau$ is a learnable temperature scaling the logits, and $b$ is a bias term. In summary, our model enables adaptation to unseen graph domains without fine-tuning through three mechanisms. First, domain-aware prompt tokens encode domain-level structural patterns, guiding the model to focus on link-indicative signals within RA. Second, positive and negative support sets provide concrete references for calibrating query edges against observed structures. Finally, the similarity-based scoring mechanism performs relative comparison instead of absolute classification, improving robustness to distribution shifts. Together, these components enable efficient and robust link inference in new graph domains.

\subsubsection{Model Training}
On the training datasets, we optimize the model by minimizing the binary cross-entropy (BCE) loss between the predicted link probabilities $\tilde{y}_q$ and the ground-truth labels $y_q \in \{0,1\}$ of query edges:

\begin{equation}
    \mathcal{L} = - \frac{1}{|\mathcal{T}|} \sum_{{e_i} \in \mathcal{T}} \Big[ y_{e_i} \log \tilde{y}_{e_i} + (1 - y_{e_i}) \log (1 - \tilde{y}_{e_i}) \Big],
    \label{eq11}
\end{equation}

where $\mathcal{T}$ denotes the set.
The framework is trained in an end-to-end episodic manner, enabling the model to capture transferable topological connection patterns that generalize robustly to unseen target domains for training-free inference. Detailed algorithmic descriptions are provided in Appendix~\ref{app:Pseudo-Code}.

\subsection{Complexity Analysis}\label{app:complex}

We analyze the computational complexity of \ourmethod, which consists of Relation Atlas (RA) construction and Prompt-guided In-context Prediction.

In RA construction, global perception is computed via $K_{gp}$-order sparse matrix propagation, while local refinement is obtained through $K_{lr}$-hop structural statistics. Both rely on sparse operations over the edge set, yielding a unified complexity of $\mathcal{O}(K_{\max}|\mathcal{E}|)$, where $K_{\max} = \max(K_{gp}, K_{lr})$ and $|\mathcal{E}|$ denotes the number of edges.
In the Prediction stage, a Transformer operates over $K$ domain prompts, $|\mathcal{S}|$ support samples, and $|\mathcal{Q}|$ query edges with hidden dimension $d$. Each query attends only to a fixed-size context composed of prompts and support samples, resulting in complexity $\mathcal{O}(|\mathcal{Q}|(K+|\mathcal{S}|)d + (K+|\mathcal{S}|)^2 d)$. Since $K$ and $|\mathcal{S}|$ are small constants, the inference cost reduces to $\mathcal{O}(|\mathcal{Q}|d)$.
Overall, the total complexity of \ourmethod is:
$\mathcal{O}(K_{\max}|\mathcal{E}| + |\mathcal{Q}|d)$.

%% file: backup/exp.tex
\label{exp}

\subsection{Experimental Setup}
\textbf{Datasets.}
Following prior work on universal link prediction, we collect graph datasets from diverse domains, including biology (\tra{Ecoli, Yeast}, \tes{Celegans})~\cite{von2002comparative,watts1998collective,zhang2018beyond}, transport (\tra{Power}, \tes{USAir})~\cite{batagelj2014pajek,watts1998collective}, web (\tra{PolBlogs, Router}, \tes{PB})~\cite{ackland2005mapping,adamic2005political,spring2002measuring}, collaboration (\tra{Physics}, \tes{CS, NS})~\cite{newman2006finding,shchur2018pitfalls}, citation (\tra{Pubmed, Citeseer}, \tes{Cora})~\cite{zhang2017weisfeiler}, and social networks (\tra{Twitch, Github}, \tes{Facebook})~\cite{rozemberczki2021multi}. 
The datasets are partitioned into two disjoint groups: a \tra{pre-training (source) set} and a \tes{testing (target) set}.
For each target graph, edges are split into 70\%, 10\%, and 20\% for context (observed edges), validation, and testing (unobserved edges), respectively. 
Detailed statistics are provided in Appendix~\ref{app:data}.

\textbf{Comparison Methods.}
We compare \ourmethod with a comprehensive set of representative baselines, including: 
(1) classical heuristic methods, such as Common Neighbors (CN)~\cite{liben2003link}, Adamic-Adar (AA)~\cite{adamic2003friends}, Resource Allocation (RA)~\cite{zhou2009predicting}, Preferential Attachment (PA)~\cite{barabasi1999emergence}, Shortest Path (SP), and Katz index~\cite{katz1953new}.; 
(2) supervised and pretraining-based link prediction models, including SEAL~\cite{zhang2018link}, GAE~\cite{kipf2016variational}, ELPH~\cite{chamberlain2022graph}, NCNC~\cite{wang2023neural}, and MPLP~\cite{dong2024pure}; 
and (3) the state-of-the-art universal link prediction model UniLP~\cite{dong2024universal}. 
Detailed descriptions of these methods are provided in Appendix~\ref{app:baseline}.

\textbf{Evaluation and Implementation.}
Following prior work, we adopt Hits@50 as the evaluation metric for link prediction, reporting the average performance over multiple runs with different random seeds to ensure reliable comparisons.
To systematically assess cross-domain transferability, parametric GNN baselines are evaluated under two settings: (1) \textit{Pretrain Only}, where pretrained models are directly applied to target graphs without any parameter updates; and (2) \textit{Pretrain \& Finetune}, where pretrained models are further adapted to target graphs using 200 sampled positive and negative edges from the context split.
In contrast, \ourmethod and UniLP perform inference on unseen target graphs in a strictly training-free manner, without requiring any additional parameter updates or fine-tuning.
Further implementation details and experimental configurations are provided in Appendix~\ref{app:imple}.

\input{tables/main_table}

\subsection{Experimental Results}
\textbf{Performance Comparison.} We evaluate \ourmethod on 7 unseen target graphs, results are reported in Table~\ref{tab:main_results}. \ding{182}~\ourmethod achieves SOTA performance, attaining the best average Hits@50 (71.69\%) and consistently outperforming all baselines, benefiting from the integration of Relation Atlas and domain-aware in-context inference for effective training-free transfer.
\ding{183}~It surpasses both heuristics and GNNs, with gains of 7.85\% over Katz and 4.83\% over SEAL, This benefits from domain-aware prompt learning and in-context prediction, which enable adaptive modeling of target-specific connectivity patterns.
\ding{184}~It also outperforms UniLP by 3.69\% on average, as Relation Atlas captures richer multi-scale structural signals (global and local), yielding stronger representations for link inference.

 \begin{wrapfigure}{r}{0.35\textwidth}
 \centering
 \vspace{-4mm}
  \includegraphics[width=0.35\textwidth]{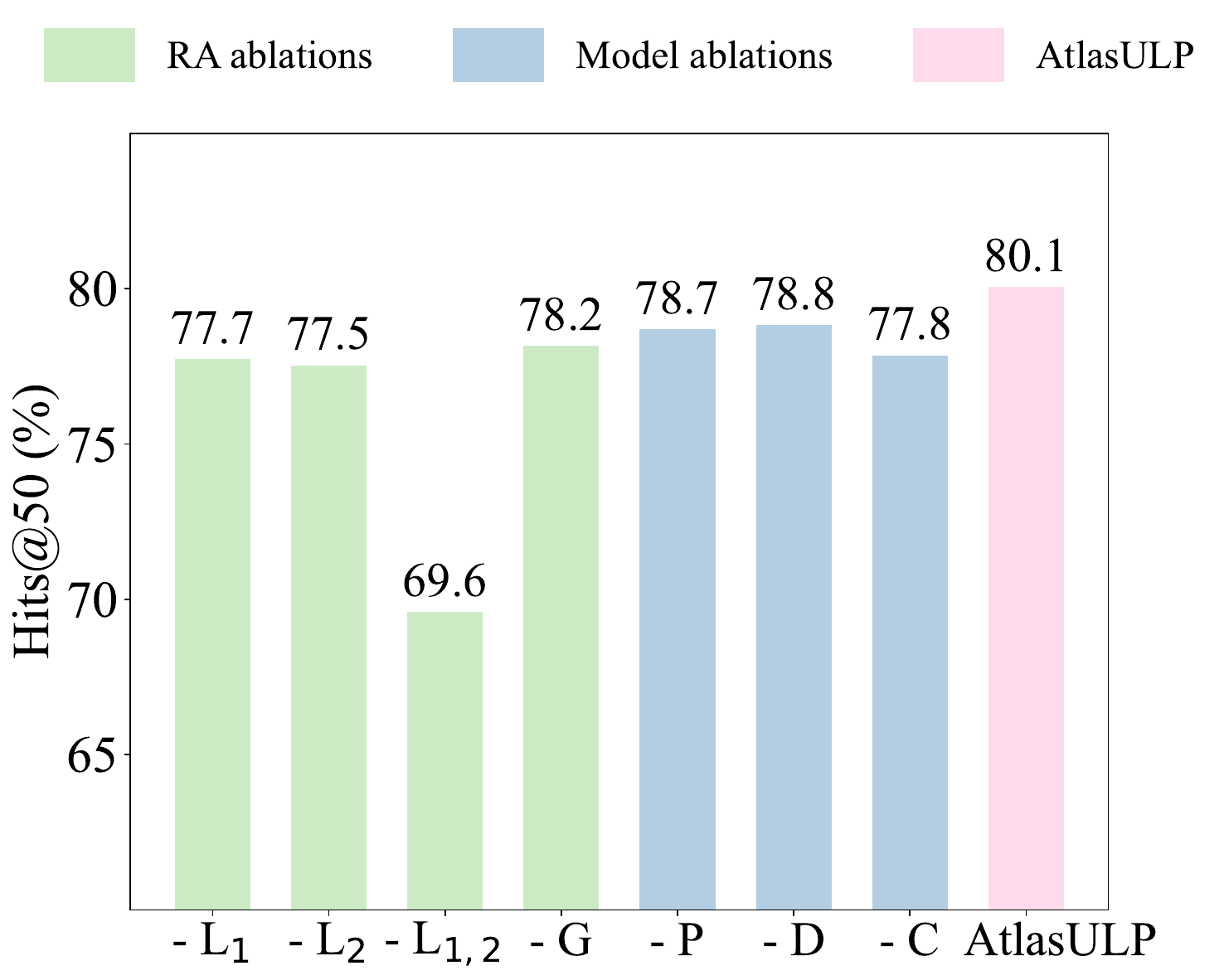}
    \vspace{-6mm}
  \caption{Ablation Studies}
  \label{fig:ablation}
    \vspace{-2mm}
\end{wrapfigure}

\textbf{Ablation Study of Relation Atlas.}
To assess the contribution of each component in the proposed Relation Atlas (RA), we conduct experiments on four datasets (Celegans, USAir, NS, and Facebook), considering four variants: removing local connectivity patterns ($-\text{L}_1$), removing structural role similarity ($-\text{L}_2$), removing the entire Local Refinement ($-\text{L}_{1,2}$), and removing Global Perception ($-\text{G}$).
All variants consistently degrade performance, indicating that each component contributes to representation quality (Fig.~\ref{fig:ablation}). In particular, removing Local Refinement ($-\text{L}_{1,2}$) leads to an average 10.5\% drop, highlighting the importance of local structural signals. These results confirm that RA effectively integrates complementary global and local information for structural modeling.

\textbf{Ablation Study of ULP Model.}
To evaluate the contribution of each component in the proposed Domain-guided Prompting ULP model, we consider three variants: removing domain-aware prompt learning ($-\text{P}$), removing deterministic structural sampling ($-\text{D}$), and removing in-context scoring ($-\text{C}$). 
Removing any component consistently degrades performance, indicating that each module plays a critical role (Fig.~\ref{fig:ablation}). Moreover, these components work in a complementary manner to enable effective cross-domain generalization.

\begin{wrapfigure}{r}{0.35\textwidth}
 \centering
  \includegraphics[width=0.35\textwidth]{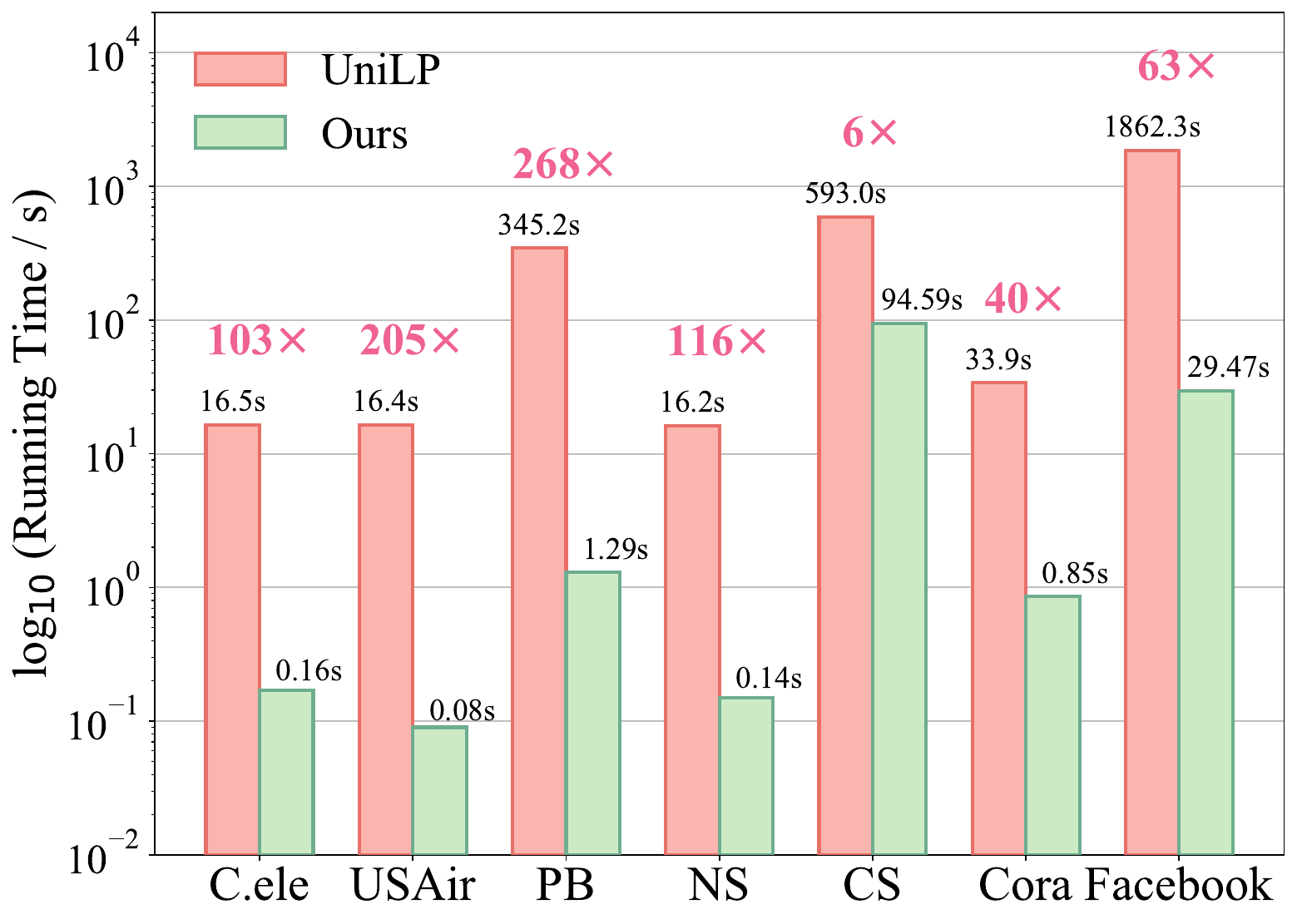}
    \vspace{-6mm}
  \caption{Time comparison}
  \label{fig:efficiency}
  \vspace{-6mm}
\end{wrapfigure}
\textbf{Efficiency Analysis}
To evaluate computational efficiency, we compare the running time of our method with UniLP on test graphs. As shown in Fig.~\ref{fig:efficiency}, Our method achieves significant speedups across graph types. For example, on Facebook, UniLP requires 1862.31s while ours takes 29.47s ($63\times$ speedup); on PB, the speedup reaches $268\times$ (345.2s vs. 1.29s).
These gains stem from our efficient unified universal representation, which eliminates the need for expensive and repeated ego-subgraph extraction required by UniLP.

\begin{figure*}[h!]
 \centering
  \includegraphics[width=0.8\textwidth]{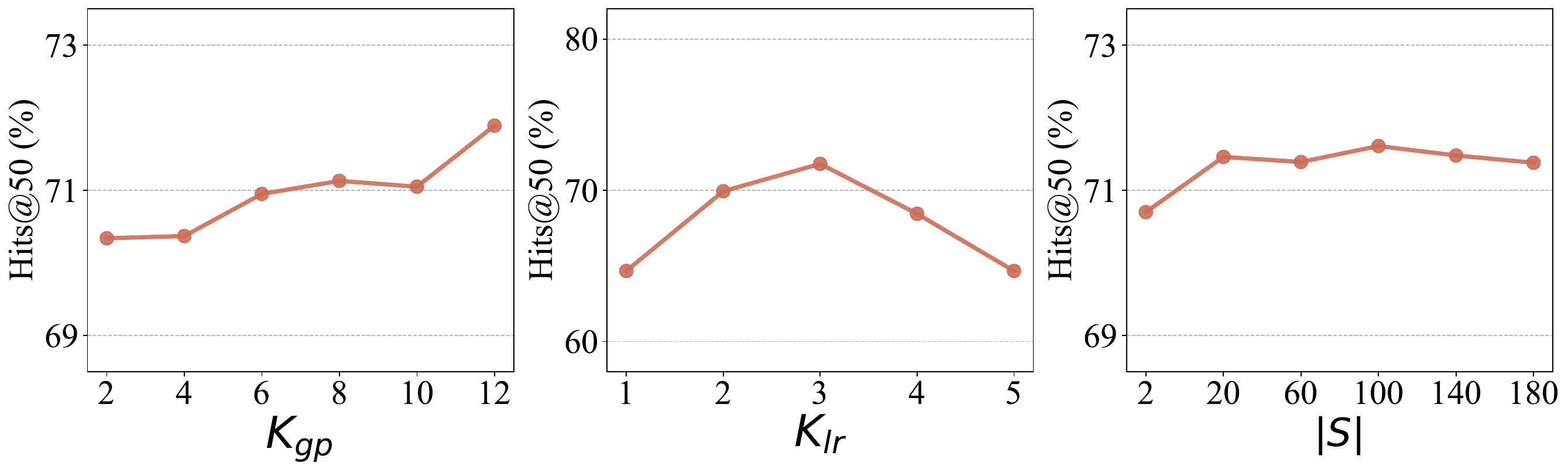}
  \caption{Parameter Sensitivity}
  \label{fig:PS}
\end{figure*}

\textbf{Parameter Sensitivity Analysis}
We investigate the sensitivity of \ourmethod to three key hyper-parameters: global perception hops $K_{gp}$, local refinement hops $K_{lr}$, and the support set size $|\mathcal{S}|$. The results are illustrated in Fig.~\ref{fig:PS}. First, the performance of our method gradually improves as $K_{gp}$ increases, as a larger $K_{gp}$ provides richer and more comprehensive global information. Second, our method achieves optimal performance when $K_{lr}$ is set to 3. This is because an excessively small $K_{lr}$ is insufficient to fully characterize local neighborhoods, whereas an overly large $K_{lr}$ introduces noisy signals due to the exponential growth of neighborhoods. Finally, the performance of our method stabilizes once $|\mathcal{S}|$ exceeds 20, demonstrating that our proposed  deterministic sampling effectively captures representative contextual references.

\begin{wrapfigure}{r}{0.35\textwidth}
 \centering
  \vspace{-3mm}
  \includegraphics[width=0.35\textwidth, trim = 115pt 70pt 110pt 140pt, clip]{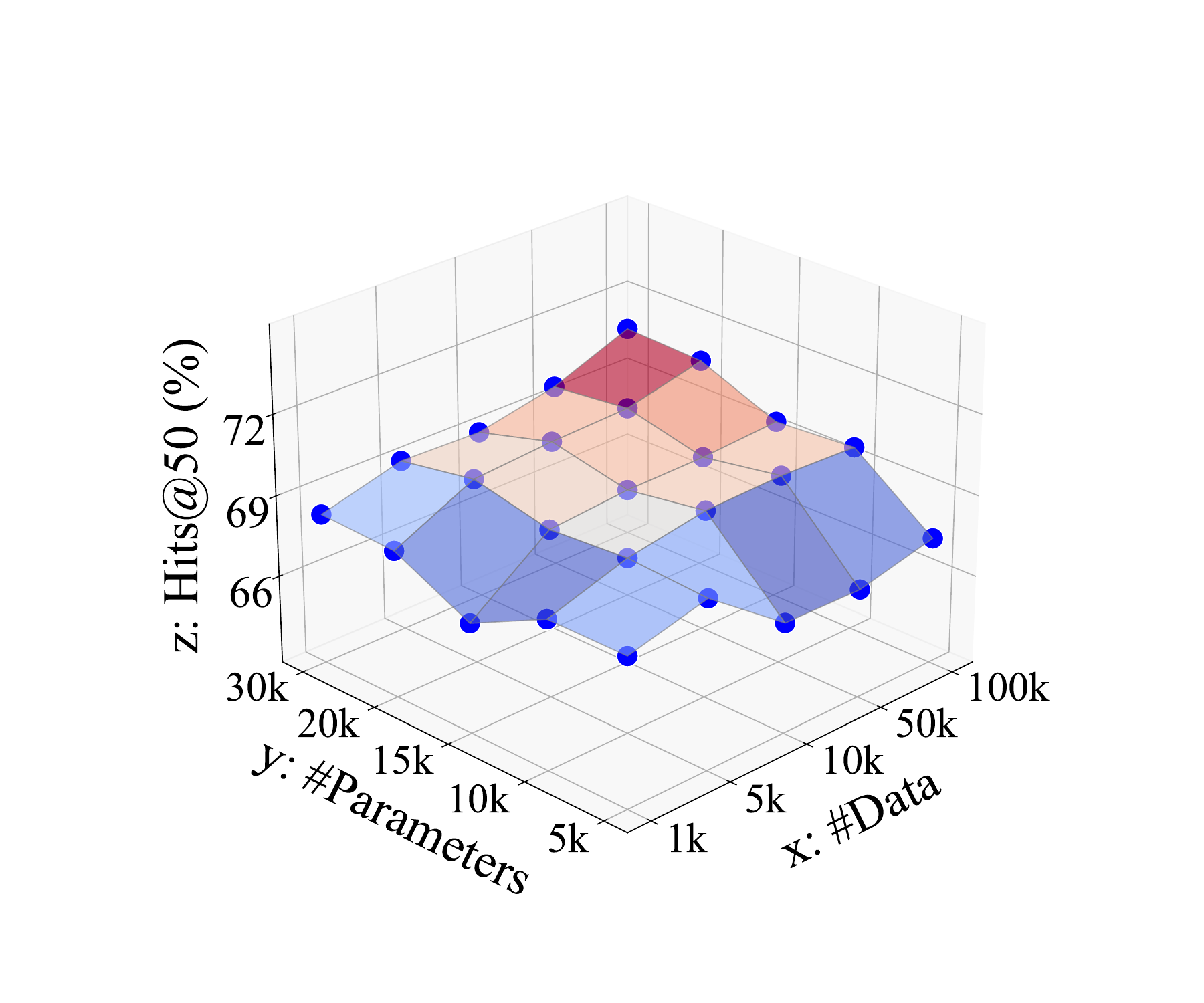}
    \vspace{-6mm}
  \caption{Scaling Law}
  \label{fig:scal}
    \vspace{-4mm}
\end{wrapfigure}
\textbf{Model Generalization Ability Analysis}
To assess the scalability of \ourmethod as a universal model, we study the relationship between its predictive performance (Hits@50) and two key factors: training data volume and model size (Fig.~\ref{fig:scal}). Performance improves monotonically with both. At larger model scales ($\sim$30k parameters), more data ($\sim$100k samples) is required to fully utilize the model capacity. The best results occur in the large-data, large-model regime (top red region), where Hits@50 exceeds 70\%. This trend aligns with standard scaling laws, demonstrating the effectiveness of \ourmethod in large-scale, cross-domain settings.

%% file: tables/main_table.tex
\begin{table*}[t]
\centering
\caption{Link prediction results on test datasets evaluated by Hits@50. The format is average score ± standard deviation. The 
top three models are colored by \textcolor{win1}{First}, \textcolor{win2}{Second}, \textcolor{win3}{Third}.}
\label{tab:main_results}
\renewcommand{\arraystretch}{1.15}
\resizebox{\textwidth}{!}{
\begin{tabular}{l c c c c c c c c}
\toprule
\textbf{Method} & \textbf{C.ele} & \textbf{USAir} & \textbf{PB} & \textbf{NS} & \textbf{CS} & \textbf{Cora} & \textbf{Facebook} & \textbf{Ave.} \\
\midrule
\rowcolor{gray!20}
\multicolumn{9}{c}{\textbf{Heuristics}} \\
\textbf{CN}   & 46.88$\pm$12.28 & 82.75$\pm$1.54 & 41.15$\pm$3.77 & 74.03$\pm$1.59 & 56.84$\pm$15.56 & 33.85$\pm$0.93 & 58.70$\pm$0.35 & 56.31 \\
\textbf{AA}   & 61.07$\pm$5.16 & 86.96$\pm$2.24 & 44.12$\pm$3.36 & 74.03$\pm$1.59 & 68.22$\pm$1.08 & 33.85$\pm$0.93 & \textcolor{win2}{\textbf{67.80$\pm$2.12}} & 62.29 \\
\textbf{RA}   & 62.80$\pm$4.84 & \textcolor{win3}{\textbf{87.27$\pm$1.89}} & 43.72$\pm$2.86 & 74.03$\pm$1.59 & 68.21$\pm$1.08 & 33.85$\pm$0.93 & \textcolor{win1}{\textbf{68.84$\pm$2.03}} & 62.67 \\
\textbf{PA}   & 43.85$\pm$4.12 & 77.69$\pm$2.29 & 28.93$\pm$1.91 & 35.35$\pm$3.01 & 6.49$\pm$0.61 & 22.09$\pm$1.52 & 12.95$\pm$0.63 & 32.48 \\
\textbf{SP}   & 0.00$\pm$0.00 & 0.00$\pm$0.00 & 0.00$\pm$0.00 & 80.00$\pm$1.11 & 41.34$\pm$35.58 & 52.97$\pm$1.53 & 0.00$\pm$0.00 & 24.90 \\
\textbf{Katz} & 58.86$\pm$6.48 & 84.64$\pm$1.86 & 44.36$\pm$3.65 & 78.96$\pm$1.35 & 66.32$\pm$5.59 & 52.97$\pm$1.53 & 60.79$\pm$0.60 & 63.84 \\
\midrule
\rowcolor{gray!20}
\multicolumn{9}{c}{\textbf{Pretrain Only}} \\
\textbf{SEAL} & 61.28$\pm$3.76 & 86.00$\pm$1.56 & 45.44$\pm$2.68 & 84.07$\pm$1.96 & 62.82$\pm$1.62 & 56.21$\pm$2.24 & 54.57$\pm$1.48 & 64.34 \\
\textbf{GAE}  & 44.71$\pm$3.39 & 76.12$\pm$2.27 & 27.56$\pm$2.34 & 15.20$\pm$2.14 & 5.08$\pm$0.48 & 24.22$\pm$1.53 & 6.65$\pm$0.57 & 28.51 \\
\textbf{ELPH} & 59.23$\pm$4.50 & 84.42$\pm$2.22 & 43.69$\pm$2.90 & 84.27$\pm$1.43 & \textcolor{win2}{\textbf{70.69$\pm$3.63}} & 56.91$\pm$1.43 & 61.80$\pm$2.46 & 65.86 \\
\textbf{NCNC} & 48.07$\pm$4.79 & 75.44$\pm$4.22 & 25.66$\pm$1.42 & 80.07$\pm$1.43 & 34.27$\pm$2.28 & 52.51$\pm$2.54 & 19.28$\pm$1.58 & 47.90 \\
\textbf{MPLP} & 56.74$\pm$5.31 & 82.94$\pm$2.30 & 47.78$\pm$2.55 & 80.33$\pm$1.54 & 24.26$\pm$1.28 & 46.71$\pm$2.25 & 48.06$\pm$2.09 & 55.26 \\
\midrule
\rowcolor{gray!20}
\multicolumn{9}{c}{\textbf{Pretrain \& Finetune}} \\
\textbf{SEAL} & \textcolor{win3}{\textbf{64.45$\pm$4.14}} & \textcolor{win2}{\textbf{88.49$\pm$2.16}} & 47.78$\pm$3.32 & 84.84$\pm$2.32 & 61.54$\pm$3.09 & \textcolor{win2}{\textbf{62.19$\pm$3.27}} & 58.70$\pm$2.78 & \textcolor{win3}{\textbf{66.86}} \\
\textbf{GAE}  & 44.71$\pm$4.07 & 74.47$\pm$2.96 & 25.92$\pm$2.64 & 18.34$\pm$2.27 & 4.95$\pm$0.44 & 25.31$\pm$1.48 & 6.11$\pm$0.39 & 28.54 \\
\textbf{ELPH} & 60.51$\pm$5.72 & 84.52$\pm$2.08 & 43.58$\pm$3.48 & \textcolor{win3}{\textbf{86.08$\pm$0.69}} & \textcolor{win1}{\textbf{71.10$\pm$3.48}} & 57.18$\pm$1.89 & 63.31$\pm$3.63 & 66.61 \\
\textbf{NCNC} & \textcolor{win3}{\textbf{64.45$\pm$5.10}} & 85.85$\pm$2.46 & 47.75$\pm$6.90 & \textcolor{win2}{\textbf{88.25$\pm$2.25}} & 58.75$\pm$5.74 & \textcolor{win3}{\textbf{60.00$\pm$2.50}} & 59.32$\pm$6.93 & 66.34 \\
\textbf{MPLP} & 62.56$\pm$4.79 & 85.08$\pm$1.54 & \textcolor{win3}{\textbf{48.01$\pm$2.94}} & 80.16$\pm$1.18 & 50.35$\pm$1.01 & 56.02$\pm$2.09 & 56.72$\pm$1.20 & 62.70 \\
\midrule
\rowcolor{gray!20}
\multicolumn{9}{c}{\textbf{Universal LP model}} \\
\textbf{UniLP}& \textcolor{win2}{\textbf{65.20$\pm$4.40}} & 85.98$\pm$2.00 & \textcolor{win2}{\textbf{48.14$\pm$2.99}} & \textcolor{win1}{\textbf{89.09$\pm$2.05}} & 64.59$\pm$2.65 & 57.50$\pm$2.40 & 65.49$\pm$2.05 & \textcolor{win2}{\textbf{68.00}} \\
\textbf{Ours}  & \textcolor{win1}{\textbf{77.30$\pm$1.20}} & \textcolor{win1}{\textbf{88.80$\pm$0.77}} & \textcolor{win1}{\textbf{48.84$\pm$0.33}} & \textcolor{win2}{\textbf{88.25$\pm$1.23}} & \textcolor{win3}{\textbf{70.32$\pm$0.51}} & \textcolor{win1}{\textbf{62.41$\pm$0.64}} & \textcolor{win3}{\textbf{65.91$\pm$1.51}} & \textcolor{win1}{\textbf{71.69}} \\
\bottomrule
\end{tabular}
}
\end{table*}

%% file: backup/conclusion.tex
In this paper, we propose a novel universal link prediction method, \ourmethod, aiming to construct a universal LP representation that comprehensively captures link-indicative information across diverse graph domains. We introduce the Relation Atlas (RA), which characterizes structural correlations between node pairs from both local and global perspectives. Built upon this, we further develop an Atlas-grounded ULP model that identifies domain-specific connectivity patterns via domain-aware prompting and enables effective adaptive transfer through prompt-based in-context prediction. Extensive experiments on graph datasets from six real-world domains demonstrate that \ourmethod achieves strong predictive performance, positive transferability, and scalable capability. The main \textbf{limitations} of \ourmethod lies in the need for manual design in constructing the Relation Atlas. Therefore, a promising \textbf{future directions} is to explore automated relational atlas learning.

%% file: backup/appendix.tex
\section{Detail of \ourmethod}
\subsection{Local Refinement Construction}
\label{app: local}
Although global perception captures global signals, it struggles with local structural interactions. Thus We introduce Local Refinement, which replaces subgraph sampling with lightweight statistical indicators to model local structure efficiently. Specifically, we model local connectivity patterns and structural role similarity using five metrics, starting with three measures of local connectivity: Hierarchical Neighborhood Overlap, Resource-Allocated Overlap, and Local Connectivity Density:

\textbf{\ding{182} Hierarchical Neighborhood Overlap (HN)} characterizes neighbor overlap across multi-hop neighborhoods. We construct the multi-hop frontier sets for nodes $v_i$ and $v_j$ and calculate the intersection size between different hop levels:

\begin{equation}
\mathbf{s}_{\text{HN}}^{(i,j)} = \text{Concat} \left( \left\{ |\mathcal{F}_i^{(a)} \cap \mathcal{F}_j^{(b)}| \;\middle|\; a, b \in \{1, \dots, K_{lr}\} \right\} \right),
\label{eq12}
\end{equation}
where $\mathcal{F}_i^{(a)}$ denotes the $a$-hop frontier set of node $v_i$, directly obtained from the non-zero entries of $\mathbf{T}_a$ in Eq.~\ref{eq1}. $K_{lr}$ denotes the maximum path order considered in local refinement. This metric captures neighborhood overlap across structural scales.

\textbf{\ding{183} Resource-Allocated Overlap(RA)} aims to differentiate the contributions of various common neighbors. Low-degree neighbors often indicate a tighter connection than high-degree hubs. Therefore, we introduce degree information to weight the overlapping region:

\begin{equation}
\mathbf{s}_{\text{RA}}^{(i,j)} = \text{Concat} \left( \left\{ \sum_{v_k \in \mathcal{F}_i^{(a)} \cap \mathcal{F}_j^{(b)}} \frac{1}{d(v_k)} \;\middle|\; a, b \in \{1, \dots, K_{lr}\} \right\} \right),
\label{eq13}
\end{equation}
where $d(v_k)$ represents the degree of node $v_k$. While HN models the quantity of common neighbors, RA distinguishes their importance. Combining both provides a comprehensive view of neighborhood overlap.

\textbf{\ding{184} Local Connectivity Density (LC)} further quantifies the connection density between cross-level neighborhoods, capturing topological compactness that cannot be expressed by overlap alone. We calculate the edge connection strength between different frontiers:

\begin{equation}
\mathbf{s}_{\text{LC}}^{(i,j)} = \text{Concat} \left( \left\{ (\mathbf{f}_i^{(a)})^\top \mathbf{A} \mathbf{f}_j^{(b)} \;\middle|\; a, b \in \{1, \dots, K_{lr}\} \right\} \right),
\label{eq14}
\end{equation}
where $\mathbf{f}_i^{(a)} \in \{0,1\}^n$ is the binary indicator vector of the frontier set $\mathcal{F}_i^{(a)}$. 
Through these three metrics, we obtain a detailed characterization of local subgraph structure. As all metrics are computed via sparse matrix operations without subgraph extraction, the design is computationally efficient. 

Next, we model structural role similarity between node pairs. Structural role reflects a node’s position in local topology (e.g., hubs or peripheral nodes). Nodes with similar roles tend to exhibit stronger connectivity, making this signal useful for link prediction. We characterize structural role similarity using two metrics: Frontier Size Product and Frontier Size Difference:

\textbf{\ding{185} Frontier Size Product (FP)} measures the product of frontier set sizes across different orders of node pairs:

\begin{equation}
\mathbf{s}_{\text{FP}}^{(i,j)} = \text{Concat} \left( \left\{ |\mathcal{F}_i^{(a)}| \cdot |\mathcal{F}_j^{(a)}| \;\middle|\; a \in \{1, \dots, K_{lr}\} \right\} \right).
\label{eq15}
\end{equation}

\textbf{\ding{186} Frontier Size Difference (FD)} measures the absolute discrepancy in neighborhood volumes of the node pair:

\begin{equation}
\mathbf{s}_{\text{FD}}^{(i,j)} = \text{Concat} \left( \left\{ \left| |\mathcal{F}_i^{(a)}| - |\mathcal{F}_j^{(a)}| \right| \;\middle|\; a \in \{1, \dots, K_{lr}\} \right\} \right).
\label{eq16}
\end{equation}

By jointly modeling these two metrics, we obtain a comprehensive characterization of structural role similarity. For example, a small FD indicates similar roles; when coupled with a high FP, it suggests symmetric hub-like nodes, which tend to exhibit higher link formation likelihood than symmetric peripheral ones.
By combining all five structural metrics, we obtain a unified representation of local connectivity patterns and structural role similarity for node pairs. The local refinement representation is defined as follows:

\begin{equation}
\mathbf{h}_{local}^{(i,j)} = \left[ \mathbf{s}_{\text{HN}}^{(i,j)} \parallel \mathbf{s}_{\text{RA}}^{(i,j)} \parallel \mathbf{s}_{\text{LC}}^{(i,j)} \parallel \mathbf{s}_{\text{FP}}^{(i,j)} \parallel \mathbf{s}_{\text{FD}}^{(i,j)} \right].
\label{eq17}
\end{equation}

Finally, by combining the global perception and local refinement representations, we obtain a comprehensive modeling of multi-scale topological signals, forming an effective universal LP representation.

\subsection{Attention Mask and Transformer}
\label{app: mask}
To enforce the inductive setting in in-context inference, we design an attention mask $\mathbf{M} \in \mathbb{R}^{L_{seq} \times L_{seq}}$ to prevent information leakage among query samples. Specifically, domain prompts and support samples are fully visible, while each query can only attend to prompts, support, and itself. The mask is defined as:

\begin{equation}
    m_{i,j} = \begin{cases} 0, & \text{if } j \le K + |\mathcal{S}^+| + |\mathcal{S}^-| \text{ or } i = j, \\ -\infty, & \text{otherwise}. \end{cases}
    \label{eq18}
\end{equation}

We then feed the input sequence $\mathbf{Z}^{(0)}$ along with the mask $\mathbf{M}$ into an $L$-layer Transformer encoder to obtain the deep contextualized representation $\mathbf{Z}^{(L)}$, where each layer can be formulated as:

\begin{equation}
    \mathbf{Z}^{(l+1)} = \text{FFN}\left( \text{Softmax}\left( \frac{\mathbf{Q}\mathbf{K}^\top}{\sqrt{d'}} + \mathbf{M} \right) \mathbf{V} \right),
    \label{eq19}
\end{equation}
where $\mathbf{Q}, \mathbf{K}, \mathbf{V}$ are the query, key, and value matrices, computed by $\mathbf{Q} = \mathbf{Z}^{(l)} \mathbf{W}_Q^{(l)}$, $\mathbf{K} = \mathbf{Z}^{(l)} \mathbf{W}_K^{(l)}$, and $\mathbf{V} = \mathbf{Z}^{(l)} \mathbf{W}_V^{(l)}$, with each $\mathbf{W}$ as learnable parameters, and $d'$ represents the dimension of the attention heads.

\section{Algorithm of \ourmethod}
\label{app:Pseudo-Code}

The algorithm pseudo-code of \ourmethod is shown in Algo.~\ref{algo:algo}.

\begin{algorithm}[htbp]
\caption{Pseudo-Code of the Proposed \ourmethod}\label{algo:algo}
\begin{algorithmic}[1] 

\REQUIRE Source-domain graphs $\{\mathcal{G}_{s}^{(1)}, \cdots, \mathcal{G}_{s}^{(N)}\}$; \\
Target-domain graph $\mathcal{G}_t$; \\
Model parameters $\Theta$.

\ENSURE Trained universal LP model with parameters $\Theta^{*}$.

\STATE \textbf{Step 1: Relation Atlas Extraction and Encoding}
\FOR{each graph $\mathcal{G}=(\mathcal{V},\mathcal{E})$}
    \STATE Compute global perception representation $\mathbf{H}_{global}$ according to Eqs.~\eqref{eq1} and~\eqref{eq2};
    \STATE Compute local refinement representation $\mathbf{H}_{local}$ according to Eqs.~\eqref{eq12}--~\eqref{eq17};
    \STATE Encoding final embedding of edges $\mathbf{h}_{e}$ according to Eq.~\eqref{eq4};
\ENDFOR

\STATE \textbf{Step 2: Train Domain-guided Prompting ULP Model}
\FOR{each training episode}
    \STATE Select background reference edges $\mathcal{C}^+_{b} = \{ e_1, \cdots, e_M \}$ and support sets $\mathcal{S}^+$ and $\mathcal{S}^-$;

    \STATE \textbf{Domain-aware Prompt Learning:}
    \STATE Encode domain-aware prompt token according to Eqs.~\eqref{eq5} and~\eqref{eq6};

    \STATE \textbf{Prompt-guided In-context Prediction:}
    \STATE Construct attention mask and the input sequence of Transformer according to Eqs.~\eqref{eq18} and~\eqref{eq8};
    \STATE Encode node representations via the Transformer according to Eqs.~\eqref{eq19};

    \STATE \textbf{Link Scoring:}
    \STATE Compute link probabilities based on support-query similarity according to Eqs.~\eqref{eq9} and~\eqref{eq10};
    \STATE Update model parameters $\Theta$ by minimizing the BCE loss according to Eq.~\eqref{eq11};
\ENDFOR\\
Return trained model parameters $\Theta^{*}$;

\STATE \textbf{Training-Free Prediction on Target Domain}
\STATE Extract relation atlas of the target graph $\mathcal{G}_t$;
\STATE Encode edge representations using the trained model with parameters $\Theta^{*}$;
\STATE Compute link probabilities for target edge.

\end{algorithmic}
\end{algorithm}

\section{Experimental Details}
\label{app:ExpDetails}
\subsection{Datasets}
\label{app:data}
We collect graph datasets from six domains, including biology, transport, web, collaboration, citation, and social networks, to evaluate the generalization of universal link prediction (ULP) models. 
Such diversity introduces heterogeneous structural patterns across graphs, posing significant challenges for transferring link prediction knowledge.
Following prior work on universal link prediction, we collect graph datasets from diverse domains, including biology (\tra{Ecoli, Yeast}, \tes{Celegans})~\cite{von2002comparative,watts1998collective,zhang2018beyond}, transport (\tra{Power}, \tes{USAir})~\cite{batagelj2014pajek,watts1998collective}, web (\tra{PolBlogs, Router}, \tes{PB})~\cite{ackland2005mapping,adamic2005political,spring2002measuring}, collaboration (\tra{Physics}, \tes{CS, NS})~\cite{newman2006finding,shchur2018pitfalls}, citation (\tra{Pubmed, Citeseer}, \tes{Cora})~\cite{zhang2017weisfeiler}, and social networks (\tra{Twitch, Github}, \tes{Facebook})~\cite{rozemberczki2021multi}. 
The datasets are partitioned into two disjoint groups: a \tra{pre-training (source) set} and a \tes{testing (target) set}.
Table~\ref{tab:dataset_statistics} summarizes the key statistics. Below we briefly describe each dataset.

\subsubsection*{Source (Pre-training) Datasets.}

\begin{itemize}
    \item[$\bullet$] \textbf{Ecoli} is a biological network where nodes represent operons and edges denote interactions.
    
    \item[$\bullet$] \textbf{Yeast} is a protein--protein interaction (PPI) network, where nodes represent proteins and edges denote physical interactions.
    
    \item[$\bullet$] \textbf{Power} is a power grid network, where nodes represent power plants or transformers and edges denote transmission lines.
    
    \item[$\bullet$] \textbf{PolBlogs} is a web network, where nodes represent weblogs and edges denote hyperlink references.
    
    \item[$\bullet$] \textbf{Router} is a web network, where nodes represent routers and edges denote physical connections.
    
    \item[$\bullet$] \textbf{Physics} is a collaboration network, where nodes represent researchers and edges denote co-authorship relations.
    
    \item[$\bullet$] \textbf{Pubmed} is a citation network, where nodes represent scientific articles and edges denote citation relationships.
    
    \item[$\bullet$] \textbf{Citeseer} is a citation network, where nodes represent publications and edges denote citation links.
    
    \item[$\bullet$] \textbf{Twitch} is a social network, where nodes represent users and edges denote interactions.
    
    \item[$\bullet$] \textbf{Github} is a social network, where nodes represent users and edges denote follower relationships.
\end{itemize}

\subsubsection*{Target (Testing) Datasets.}

\begin{itemize}
    \item[$\bullet$] \textbf{Celegans} is a biological network, where nodes represent neurons and edges denote neural connections.
    
    \item[$\bullet$] \textbf{USAir} is a transportation network, where nodes represent airports and edges denote flight routes.
    
    \item[$\bullet$] \textbf{PB} is a web network, where nodes represent blogs and edges denote hyperlink references.
    
    \item[$\bullet$] \textbf{CS} is a collaboration network, where nodes represent authors and edges denote co-authorship relations.
    
    \item[$\bullet$] \textbf{NS} is a collaboration network, where nodes represent researchers and edges denote co-authorship relations.
    
    \item[$\bullet$] \textbf{Cora} is a citation network, where nodes represent publications and edges denote citation links.
    
    \item[$\bullet$] \textbf{Facebook} is a social network, where nodes represent users and edges denote interactions.
\end{itemize}

\begin{table*}[t]
\centering
\caption{Statistics of the graph datasets from diverse domains.}
\label{tab:dataset_statistics}
\renewcommand{\arraystretch}{1.2}
\resizebox{\textwidth}{!}{
\begin{tabular}{l c c | c c c c c c}
\toprule
\textbf{Dataset} & \textbf{Pretrain} & \textbf{Test} & \textbf{\# Nodes} & \textbf{\# Edges} & \textbf{Avg. node deg.} & \textbf{Std. node deg.} & \textbf{Max. node deg.} & \textbf{Density} \\
\midrule
\rowcolor{gray!20}
\multicolumn{9}{c}{\textbf{Biology}} \\
Ecoli    & \checkmark & - & 1805 & 29320 & 16.24 & 48.38 & 1030 & 1.8009\% \\
Yeast    & \checkmark & - & 2375 & 23386 & 9.85 & 15.5 & 118 & 0.8295\% \\
Celegans & - & \checkmark & 297 & 4296 & 14.46 & 12.97 & 134 & 9.7734\% \\
\midrule
\rowcolor{gray!20}
\multicolumn{9}{c}{\textbf{Transport}} \\
Power & \checkmark & - & 4941 & 13188 & 2.67 & 1.79 & 19 & 0.1081\% \\
USAir & - & \checkmark & 332 & 4252 & 12.81 & 20.13 & 139 & 7.7385\% \\
\midrule
\rowcolor{gray!20}
\multicolumn{9}{c}{\textbf{Web}} \\
PolBlogs & \checkmark & - & 1490 & 19025 & 12.77 & 20.73 & 256 & 1.7150\% \\
Router   & \checkmark & - & 5022 & 12516 & 2.49 & 5.29 & 106 & 0.0993\% \\
PB       & - & \checkmark & 1222 & 33428 & 27.36 & 38.42 & 351 & 4.4808\% \\
\midrule
\rowcolor{gray!20}
\multicolumn{9}{c}{\textbf{Collaboration}} \\
Physics & \checkmark & - & 34493 & 495924 & 14.38 & 15.57 & 382 & 0.0834\% \\
CS      & - & \checkmark & 18333 & 163788 & 8.93 & 9.11 & 136 & 0.0975\% \\
NS      & - & \checkmark & 1589 & 5484 & 3.45 & 3.47 & 34 & 0.4347\% \\
\midrule
\rowcolor{gray!20}
\multicolumn{9}{c}{\textbf{Citation}} \\
Pubmed   & \checkmark & - & 19717 & 88648 & 4.5 & 7.43 & 171 & 0.0456\% \\
Citeseer & \checkmark & - & 3327 & 9104 & 2.74 & 3.38 & 99 & 0.1645\% \\
Cora     & - & \checkmark & 2708 & 10556 & 3.9 & 5.23 & 168 & 0.2880\% \\
\midrule
\rowcolor{gray!20}
\multicolumn{9}{c}{\textbf{Social}} \\
Twitch   & \checkmark & - & 34118 & 429113 & 12.58 & 35.88 & 1489 & 0.0737\% \\
Github   & \checkmark & - & 37700 & 289003 & 7.67 & 46.59 & 6809 & 0.0407\% \\
Facebook & - & \checkmark & 22470 & 171002 & 7.61 & 15.26 & 472 & 0.0677\% \\
\bottomrule
\end{tabular}
}
\end{table*}

\subsection{Comparison Methods}
\label{app:baseline}

\subsubsection*{Classical Heuristic Methods}
\begin{itemize}
    \item[$\bullet$] \textbf{CN}~\cite{liben2003link} (Common Neighbors) measures link likelihood based on the number of shared neighbors between two nodes.
    
    \item[$\bullet$] \textbf{AA}~\cite{adamic2003friends} (Adamic-Adar) refines CN by assigning higher weights to low-degree common neighbors.
    
    \item[$\bullet$] \textbf{RA}~\cite{zhou2009predicting} (Resource Allocation) is similar to AA but uses a different normalization scheme based on node degrees.
    
    \item[$\bullet$] \textbf{PA}~\cite{barabasi1999emergence} (Preferential Attachment) estimates link likelihood based on the product of node degrees.
    
    \item[$\bullet$] \textbf{SP} (Shortest Path) measures node proximity using the shortest path distance.
    
    \item[$\bullet$] \textbf{Katz}~\cite{katz1953new} sums over all paths between two nodes with exponential decay on longer paths.
\end{itemize}

\subsubsection*{Supervised and Pretraining-based Models}
\begin{itemize}
    \item[$\bullet$] \textbf{GAE}~\cite{kipf2016variational} is a graph auto-encoder that learns node embeddings and reconstructs graph structure for link prediction.
    
    \item[$\bullet$] \textbf{SEAL}~\cite{zhang2018link} is a subgraph-based method that extracts enclosing subgraphs and applies GNNs for link prediction.
    
    \item[$\bullet$] \textbf{ELPH}~\cite{chamberlain2022graph} captures structural features via efficient message passing for scalable link prediction.
    
    \item[$\bullet$] \textbf{NCNC}~\cite{wang2023neural} models common neighbors using neural architectures to enhance link prediction.
    
    \item[$\bullet$] \textbf{MPLP}~\cite{dong2024pure} leverages message passing to estimate structural patterns for link prediction.
\end{itemize}

\subsubsection*{Universal Link Prediction Methods}
\begin{itemize}
    \item[$\bullet$] \textbf{UniLP}~\cite{dong2024universal} is a universal link prediction framework that constructs structural representations via DRNL and performs in-context inference for cross-domain prediction.
\end{itemize}

\subsection{Evaluation and Implementation}
\label{app:imple}

\subsubsection*{Evaluation Protocol}

To evaluate the cross-domain effectiveness of \ourmethod under the universal link prediction (ULP) setting, we split all graph datasets into two disjoint groups, i.e., a pre-training (source) set and a testing (target) set (see Table~\ref{tab:dataset_statistics}). A single model is trained on the combined pre-training datasets and evaluated on all target datasets. 
To mitigate the imbalance caused by varying dataset sizes, we adopt a balanced episodic sampling strategy during pre-training. Specifically, we uniformly sample an equal number of edges from each source dataset per epoch, ensuring balanced contributions across different graph domains.

During pre-training, for each query link, we first sample 1000 candidate links from the corresponding source dataset as background reference examples $\mathcal{C}_b$. From $\mathcal{C}_b$, we further select 64 links as in-context samples.
For each target dataset, we follow a standard LP split strategy, where edges are divided into 70\%/10\%/20\% for training (observed edges), validation, and testing (unobserved edges), respectively. During inference, 1000 background reference links are sampled from each target dataset, from which 32 in-context samples are selected for prediction.
We report Hits@50 as the evaluation metric. Hits@50 measures the proportion of true positive links ranked within the top 50 predictions among all candidate links, reflecting the ranking quality of the model. Our experiments are conducted on a single NVIDIA RTX 4000 Ada GPU using PyTorch 2.7.1 with Python 3.10.

\subsubsection*{Evaluation of Baselines}

We evaluate all baseline methods under two transfer settings to ensure a fair comparison:

\textbf{Pretrain Only}: models are trained on the combined pre-training datasets and directly evaluated on each target dataset without adaptation.

\textbf{Pretrain \& Finetune}: after pretraining, models are further fine-tuned on each target dataset using 200 sampled positive and negative links.

For GAE and NCNC, which require initial node features, we use a 32-dimensional all-one vector as input features. All other methods can directly handle non-attributed graphs without additional node features.
All methods follow the same data splits and evaluation protocol to ensure a consistent and fair cross-domain comparison.

\section{Broader Impact}
\label{app: Broader Impact}
As a foundational framework for Universal Link Prediction, \ourmethod has potential positive societal impacts, such as accelerating discoveries in biological networks (e.g., protein-protein interactions) and improving the efficiency of large-scale recommendation systems.

We note that, in some applications, link prediction models may also raise potential privacy considerations when applied to sensitive social networks, where inferred connections could reveal latent relationships. Appropriate safeguards should be considered in such scenarios.

%% file: checklist.tex
\section*{NeurIPS Paper Checklist}

\begin{enumerate}

\item {\bf Claims}
    \item[] Question: Do the main claims made in the abstract and introduction accurately reflect the paper's contributions and scope?
    \item[] Answer: \answerYes{} 
    \item[] Justification: We clearly state our contributions in the abstract and introduction, and successfully validate them through extensive experiments in the subsequent sections.

\item {\bf Limitations}
    \item[] Question: Does the paper discuss the limitations of the work performed by the authors?
    \item[] Answer: \answerYes{} 
    \item[] Justification: We explicitly discuss the limitations of our work, specifically the reliance on manual design for constructing the Relation Atlas, in the Conclusion section.

\item {\bf Theory assumptions and proofs}
    \item[] Question: For each theoretical result, does the paper provide the full set of assumptions and a complete (and correct) proof?
    \item[] Answer: \answerNA{} 
    \item[] Justification: The primary contribution of this paper is an empirical methodology rather than formal mathematical theorems, though a rigorous computational complexity analysis is provided in Appendix~\ref{app:complex}.

    \item {\bf Experimental result reproducibility}
    \item[] Question: Does the paper fully disclose all the information needed to reproduce the main experimental results of the paper to the extent that it affects the main claims and/or conclusions of the paper (regardless of whether the code and data are provided or not)?
    \item[] Answer: \answerYes{} 
    \item[] Justification: All experimental details, including dataset splits, training protocols, and hyperparameters, are described in Appendix~\ref{app:ExpDetails}.

\item {\bf Open access to data and code}
    \item[] Question: Does the paper provide open access to the data and code, with sufficient instructions to faithfully reproduce the main experimental results, as described in supplemental material?
    \item[] Answer: \answerYes{} 
    \item[] Justification: We provide code and data access instructions in the supplementary material to ensure reproducibility.

\item {\bf Experimental setting/details}
    \item[] Question: Does the paper specify all the training and test details (e.g., data splits, hyperparameters, how they were chosen, type of optimizer) necessary to understand the results?
    \item[] Answer: \answerYes{} 
    \item[] Justification: We provide detailed experimental settings, including data splits, sampling strategies, and hyper-parameters in Appendix~\ref{app:ExpDetails}.

\item {\bf Experiment statistical significance}
    \item[] Question: Does the paper report error bars suitably and correctly defined or other appropriate information about the statistical significance of the experiments?
    \item[] Answer: \answerYes{} 
    \item[] Justification: We report the mean and standard deviation over 5 independent runs with different random seeds, where the standard deviation reflects the variability across runs.

\item {\bf Experiments compute resources}
    \item[] Question: For each experiment, does the paper provide sufficient information on the computer resources (type of compute workers, memory, time of execution) needed to reproduce the experiments?
    \item[] Answer: \answerYes{} 
    \item[] Justification: We describe the computational setup, including GPU usage and training configurations, in Appendix~\ref{app:imple}.
    
\item {\bf Code of ethics}
    \item[] Question: Does the research conducted in the paper conform, in every respect, with the NeurIPS Code of Ethics \url{https://neurips.cc/public/EthicsGuidelines}?
    \item[] Answer: \answerYes{} 
    \item[] Justification: This research conforms to the NeurIPS Code of Ethics in every respect.

\item {\bf Broader impacts}
    \item[] Question: Does the paper discuss both potential positive societal impacts and negative societal impacts of the work performed?
    \item[] Answer: \answerYes{} 
    \item[] Justification: We discuss the potential positive applications and negative societal impacts (e.g., privacy concerns in social networks) of universal link prediction in the Appendix~\ref{app: Broader Impact}.
    \begin{itemize}
        \item The answer \answerNA{} means that there is no societal impact of the work performed.
        \item If the authors answer \answerNA{} or \answerNo, they should explain why their work has no societal impact or why the paper does not address societal impact.
        \item Examples of negative societal impacts include potential malicious or unintended uses (e.g., disinformation, generating fake profiles, surveillance), fairness considerations (e.g., deployment of technologies that could make decisions that unfairly impact specific groups), privacy considerations, and security considerations.
        \item The conference expects that many papers will be foundational research and not tied to particular applications, let alone deployments. However, if there is a direct path to any negative applications, the authors should point it out. For example, it is legitimate to point out that an improvement in the quality of generative models could be used to generate Deepfakes for disinformation. On the other hand, it is not needed to point out that a generic algorithm for optimizing neural networks could enable people to train models that generate Deepfakes faster.
        \item The authors should consider possible harms that could arise when the technology is being used as intended and functioning correctly, harms that could arise when the technology is being used as intended but gives incorrect results, and harms following from (intentional or unintentional) misuse of the technology.
        \item If there are negative societal impacts, the authors could also discuss possible mitigation strategies (e.g., gated release of models, providing defenses in addition to attacks, mechanisms for monitoring misuse, mechanisms to monitor how a system learns from feedback over time, improving the efficiency and accessibility of ML).
    \end{itemize}
    
\item {\bf Safeguards}
    \item[] Question: Does the paper describe safeguards that have been put in place for responsible release of data or models that have a high risk for misuse (e.g., pre-trained language models, image generators, or scraped datasets)?
    \item[] Answer: \answerNA{} 
    \item[] Justification: The proposed method does not involve high-risk data or model release.

\item {\bf Licenses for existing assets}
    \item[] Question: Are the creators or original owners of assets (e.g., code, data, models), used in the paper, properly credited and are the license and terms of use explicitly mentioned and properly respected?
    \item[] Answer: \answerYes{} 
    \item[] Justification: All datasets and resources used are properly cited and follow their respective usage licenses.

\item {\bf New assets}
    \item[] Question: Are new assets introduced in the paper well documented and is the documentation provided alongside the assets?
    \item[] Answer: \answerNA{} 
    \item[] Justification: The paper evaluates the proposed model on existing public benchmark graph datasets and does not introduce new data assets.

\item {\bf Crowdsourcing and research with human subjects}
    \item[] Question: For crowdsourcing experiments and research with human subjects, does the paper include the full text of instructions given to participants and screenshots, if applicable, as well as details about compensation (if any)? 
    \item[] Answer: \answerNA{} 
    \item[] Justification: This research does not involve crowdsourcing, data labeling by human workers, or research with human subjects.

\item {\bf Institutional review board (IRB) approvals or equivalent for research with human subjects}
    \item[] Question: Does the paper describe potential risks incurred by study participants, whether such risks were disclosed to the subjects, and whether Institutional Review Board (IRB) approvals (or an equivalent approval/review based on the requirements of your country or institution) were obtained?
    \item[] Answer: \answerNA{} 
    \item[] Justification: This research does not involve human subjects, hence IRB approval is not applicable.

\item {\bf Declaration of LLM usage}
    \item[] Question: Does the paper describe the usage of LLMs if it is an important, original, or non-standard component of the core methods in this research? Note that if the LLM is used only for writing, editing, or formatting purposes and does \emph{not} impact the core methodology, scientific rigor, or originality of the research, declaration is not required.
    \item[] Answer: \answerNA{} 
    \item[] Justification: LLMs are not used as part of the proposed method.

\end{enumerate}